# Solve paint color effect prediction problem in trajectory optimization of spray-painting robot using artificial neural network inspired by the Kubelka–Munk model

Hexiang Wang[a], Zhiyuan Bi[a], Zhen Cheng[a], Xinru Li[a], Jiake Zhu[a], Liyuan Jiang[a], Hao Li[a,b*], Shizhou Lu[a,b*]

a *school of Mechanical, Electrical & Information Engineering, Shandong University, Weihai 264209, China*
b *Weihai Research Institute of Industrial Technology, Shandong University, Weihai 264209, China*
\* *Corresponding author: lushizhou@sdu.edu.cn (S.Lu)  yustlh@sdu.edu.cn (H.Li)*

**Abstract**

Currently, the spray-painting robot trajectory planning technology aiming at spray painting quality mainly applies to single-color spraying. Conventional methods of optimizing the spray gun trajectory based on simulated thickness can only qualitatively reflect the color distribution, and can not simulate the color effect of spray painting at the pixel level. Therefore, it is not possible to accurately control the area covered by the color and the gradation of the edges of the area, and it is also difficult to deal with the situation where multiple colors of paint are sprayed in combination. To solve the above problems, this paper is inspired by the Kubelka-Munk model and combines the 3D machine vision method and artificial neural network to propose a spray painting color effect prediction method. The method is enabled to predict the execution effect of the spray gun trajectory with pixel-level accuracy from the dimension of the surface color of the workpiece after spray painting. On this basis, the method can be used to replace the traditional thickness simulation method to establish the objective function of the spray gun trajectory optimization problem, and thus solve the difficult problem of spray gun trajectory optimization for multi-color paint combination spraying. In this paper, the mathematical model of the spray painting color effect prediction problem is first determined through the analysis of the Kubelka-Munk paint film color rendering model, and at the same time, the spray painting color effect dataset is established with the help of the depth camera and point cloud processing algorithm. After that, the multilayer perceptron model was improved with the help of gating and residual structure and was used for the color prediction task. To verify the practical application value of the spray painting color effect prediction model, the paper applies the model to the optimization of the spray gun trajectory with the actual workpiece. Experiments show that, in terms of prediction effect, the mean relative error of this proposed method is 7.743% and 6.814% for the test set of single-color base surface and gradient color base surface, respectively, which achieves accurate color effect prediction for a given spray gun trajectory. In terms of practical application, the method in this paper is successfully used to construct the spray gun trajectory optimization objective function, which effectively realizes the spray gun trajectory optimization.

Keywords:  Kubelka–Munk model, Artificial neural networks, Prediction of spray painting color effect, Spray-painting robot, Gun trajectory optimization

## 1. Introduction and motivation

The use of robots to automate industrial spray painting is a technology that has developed rapidly in recent years [1-3]. The use of robots to automate spray painting can be adapted to different workpiece shapes, which is particularly suitable for flexible production scenarios. In the field of robotic automatic spray painting technology, in addition to the construction and technical parameters of the spray gun, the trajectory of the robot is a key factor that affects the quality and effect of spraying [4]. When the gun construction and parameters are determined, the prediction of the spraying effect is a key step in planning the gun trajectory, which adjusts and optimizes the robot trajectory (i.e., gun trajectory) through the corresponding spraying effect prediction of the robot trajectory (i.e., gun trajectory) [5-7], and Fig. 1 shows a common flow of the optimization of the gun trajectory. Therefore, an effective spray painting color effect prediction method is important for spraying robot trajectory planning.

Currently, there are more models for predicting the paint spraying effect and producing or optimizing the gun trajectory from the perspective of thickness. Saul Nieto Bastida et al. [4], from the formula method of thickness simulation, expressed the CAD model of the workpiece through the Triangular approximation and established the optimization model of the gun trajectory by using the thickness of the simulation as the basis for evaluating the merits of the trajectory. Daniel Hegels et al. [8] also used the formula method for thickness simulation and optimized the trajectory by post-processing method based on raster trajectory. Daniel Gleeson et al. [9] combined the formula method and spline fitting to achieve the spraying simulation and generated the spraying trajectory by the pseudo projection method for a point. In addition, deep learning has also been used for spraying trajectory generation, and Tiboni et al. [10] built an end-to-end robot spraying trajectory planning model based on the PointNet (a point cloud processing deep learning model). However, in multi-color spraying scenarios (e.g., the painting scenario of a lure bait shown in Fig. 2) it is often necessary to have multiple paints coexisting on the painted surface, especially when one color of paint is partially covered on top of another, i.e., the mutual coverage of paints brings about the phenomenon of color gradient. For this phenomenon, it is difficult to accurately obtain the distribution of color on the sprayed workpiece by conventional methods, and it is also difficult to accurately obtain the location of the transition region and the rate of color change. If the paint film thickness is still used as the basis for evaluating the trajectory, it is necessary to accurately know the correspondence between the paint film thickness and the color under different base colors, which is not available in the traditional method. Therefore, it is no longer applicable to describe the spraying effect only from the perspective of paint film thickness, but rather from the perspective of pixel-level color.

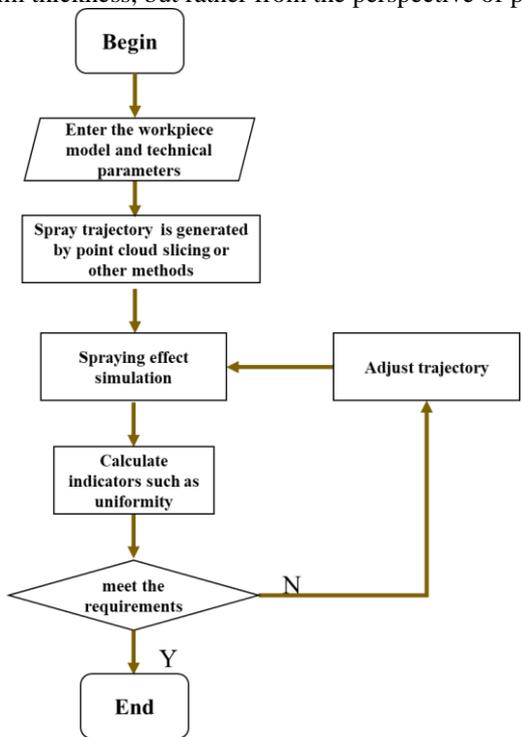

Fig. 1 A typical trajectory optimization system for a spray-painting robot

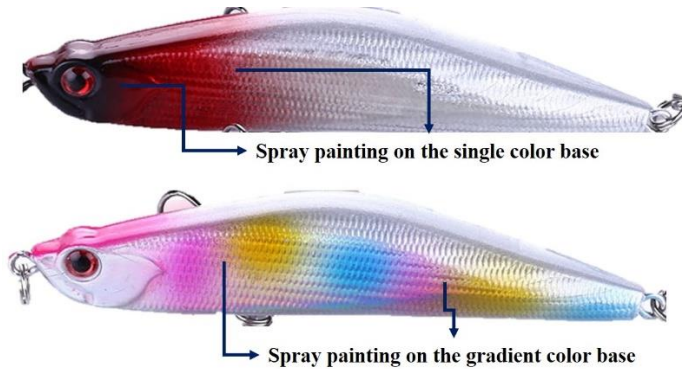

Fig. 2 The phenomenon of the color gradient of lure bait (this color gradient is achieved by spray painting)

Although there are fewer studies on paint color gradation during multi-color spraying in the field of spray painting robots, there are some techniques available in the field of digital painting. The spray gun tools of software such as Procreate and Rebelle [11] can simulate the effect of paint spraying better. There are two ways to simulate the effect of color blending, the first is based on the color blending [12] method, and the second is based on the Kubelka-Munk (K-M) model [13]. The advantage of the color blending method is that it is easy to program and implement, but the disadvantage is that it is more difficult to determine the area and speed of the color gradient (because the color gradient occurs only where the paint film thickness is thin). The advantage of the Kubelka-Munk model method is that it is a clear mechanism and the theory is reliable, but the disadvantage is that it requires expensive spectrophotometers for the experimental measurements [14] and the need for precise measurements of paint film thickness.

To overcome the disadvantages of color blending and the Kubelka-Munk model, an artificial neural network technique is introduced in this paper. A fully connected neural network (multilayer perceptron) [15] effectively establishes regression relationships between independent and dependent variables using deep learning. Neo, P. K. et al. [16] used a multilayer perceptron to predict the color of extruded thermoplastic resins, and Kose, C. et al. [17] used a multilayer perceptron to predict the final color of a white stone-reinforced ceramic restoration. In this paper, we draw on the above studies and adopt a machine-learning approach with a depth camera and a modified multilayer perceptron to realize model data acquisition and processing. The method balances prediction effectiveness and cost and can be practically applied to the field of spray-painting robots.

The purpose of this paper is to propose a spray-painting color effect prediction method for gun trajectories at the pixel-level color aspect, which in turn improves the performance of gun trajectory optimization in multi-color spraying scenarios. Focusing on this purpose, this paper analyzes the limitations of the traditional methods in Section 1 and points out that the optimization objective function should be constructed from the perspective of pixel-level color effect prediction instead of the traditional paint film thickness simulation method. Section 2 of this paper first analyzes the properties of the paint color prediction problem and then collects data from the actual painting process and processes them based on the point cloud method to construct a dataset that reacts to the color change of the painted surface during the spray painting process. Section 3 of this paper establishes the spray painting color effect prediction model and details the improvements of this model compared to the traditional multilayer perceptron model. In subsections 1-4 in Section 4 of this paper, the qualitative and quantitative results of the spray painting color effect prediction model are tested and compared with the unimproved model and other machine-learning models to demonstrate the performance of the model. Based on this, a case study applying the spray painting color effect prediction model proposed in this paper to spray gun trajectory optimization is presented in Section 4, subsection 5. The case study uses the model proposed in this paper to construct the optimization objective function, which proves the usefulness of the model.

## 2. Data collection and processing

### 2.1. Problem analysis

According to the Kubelka-Munk (K-M) model [18], the relationship between paint reflectance and wavelength and thickness can be expressed as Eq.1

$$R(\lambda, X) = \frac{1 - R_g(a - b\cosh(bSX))}{a - R_g + b\cosh(bSX)} \quad (1)$$

Where $a = 1 + K/S, b = \sqrt{a^2 - 1}$, and $R(\lambda, X), R_g = R_g(\lambda), X, K = K(\lambda), S = S(\lambda)$ are the translucent reflectance, the background reflectance, the thickness of the layer, the absorption coefficient, and the scattering coefficient, respectively.

The color of a formulated paint (i.e., when K, S is approximately fixed at the same wavelength) formed after spray painting is mainly related to the color of the surface before spray painting (i.e., $R_g(\lambda)$) and the film thickness formed by spraying (X). Accordingly, it is possible to use the simulation results of the workpiece color and paint thickness before spray painting to predict the color after spray painting. This is a multivariate regression problem. For pixel-level color prediction, the independent variables are the metric value (e.g., the value in RGB space) of the color at the point (of the pixel) before spray painting, the film thickness, and the paint color, and the dependent variable is the metric value of the color at the point (of the pixel) after spray painting.

The actual paint film thickness needs to be measured after spray painting, which is not in line with the original idea of predicting the effect of spray painting color. A natural idea is that since the paint film thickness simulation method [19] predicts the paint film thickness at each point on the surface after spray painting, the thickness obtained from the simulation can be used instead of the actual measured thickness. In this way, it is possible to predict the paint film thickness at each point on the surface based on the gun trajectory before spray painting. Afterwards, the color of each point is predicted based on the simulated thickness of the paint film. This enables pixel-level prediction of the spray painting color effect based on the gun trajectory. The formula is written as Eq.2

$$^{painted}color_{norm}(j) = Model\left(^{unpainted}color_{norm}(j), ^{unpainted}thick(j), class(j)\right) \quad (2)$$

where $^{painted}color_{norm}(j)$ denotes the normalized color of the j-th pixel (corresponding to the position on the surface) after spray painting, $^{unpainted}color_{norm}(j)$ denotes the normalized color of the j-th pixel before spray painting, $^{unpainted}thick(j)$ denotes the simulated thickness of the j-th pixel based on the gun trajectory, and $class(j)$ denotes the color of the paint corresponding to the j-th pixel sprayed, for all the areas covered by the paint present. $Model$ denotes an input-output correspondence.

However, there is a problem that the simulated thickness of the paint film is not always accurate, thus affecting the color prediction. To solve this problem, a method is needed to predict the simulated thickness of the paint film by simply having a correspondence between the simulated thickness of the paint film and the actual thickness when the equivalence relationship can not be satisfied. Because there is a correspondence between the simulated thickness and the actual thickness at a certain point, as well as a correspondence between the actual thickness and the color after spray painting, the correspondence between the simulated thickness and the color after spray painting exists, but it is not convenient to express it directly. At this time, it can be considered to start from the data collected during the spraying of paint, and mine the relationship from the data. In this paper, we design a complete set of methods to collect the data of which $^{painted}color_{norm}(j)$, $^{unpainted}color_{norm}(j)$, $^{unpainted}thick(j)$, $class(j)$ during the spraying process.

Before collecting the data, the following reasonable assumptions in the experimental and production environments are carried out: Assumption 1 is that the light stabilization does not change; the reason for the assumption is that both the experimental and production environments can be approximated. Assumption 2 is the absolute uniformity of light at all points of the workpiece (same intensity); reason for assumption is that it can be approximated by ring lighting. Assumption 3 is that no chemical reaction occurs when the overlapping of paint coatings of different colors occurs. Reason for assumption is thst in the production environment, generally the lower layer of paint dries before spray painting the upper layer of paint, so it can be approximated to realize.

## 2.2. Construction of data sets

This paper investigates the prediction of the color effect of spray painting, taking water-based metallic antirust paint for experiments, and a robotic spray gun composite system to execute the spraying. The system consists of a set of composite mechanisms that can accurately control the switch of the spray gun, a Realsense-D435i depth camera that can carry out three-dimensional reconstruction of the sprayed workpiece, a robotic arm that can drive the spray gun to execute the trajectory, as well as several mechanisms used to carry the experimental workpiece. The system is shown in Fig. 3

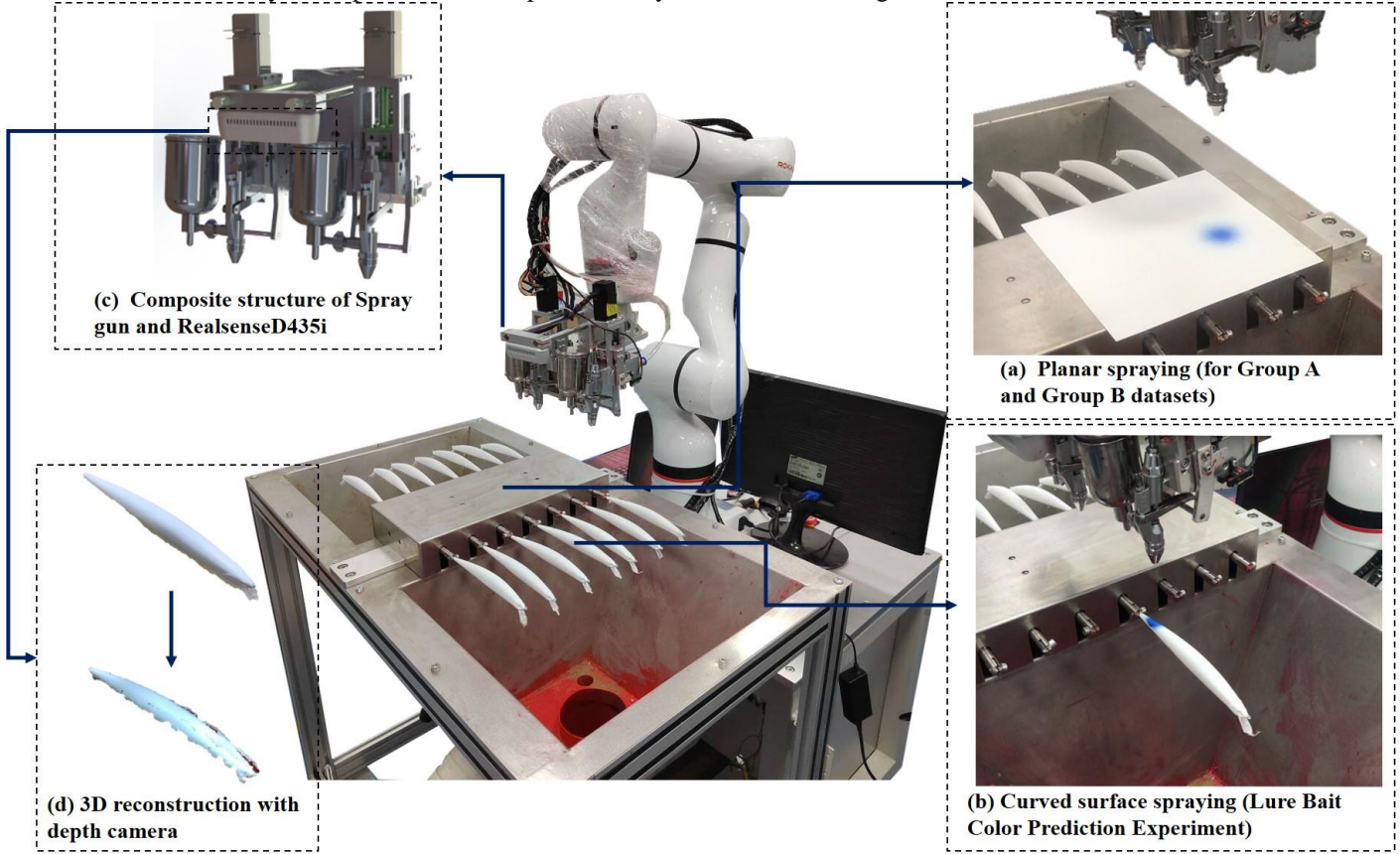

**Fig. 3 Spraying experimental system (a. Experimental method of spraying flat surfaces b. Experimental method of spraying curved surfaces c. Gun-camera composite mechanism d. Three-dimensional reconstruction of workpieces)**

In the following, the color of the experimental surface before spray painting is referred to as the base color, and the color after spray painting is referred to as the painted color.

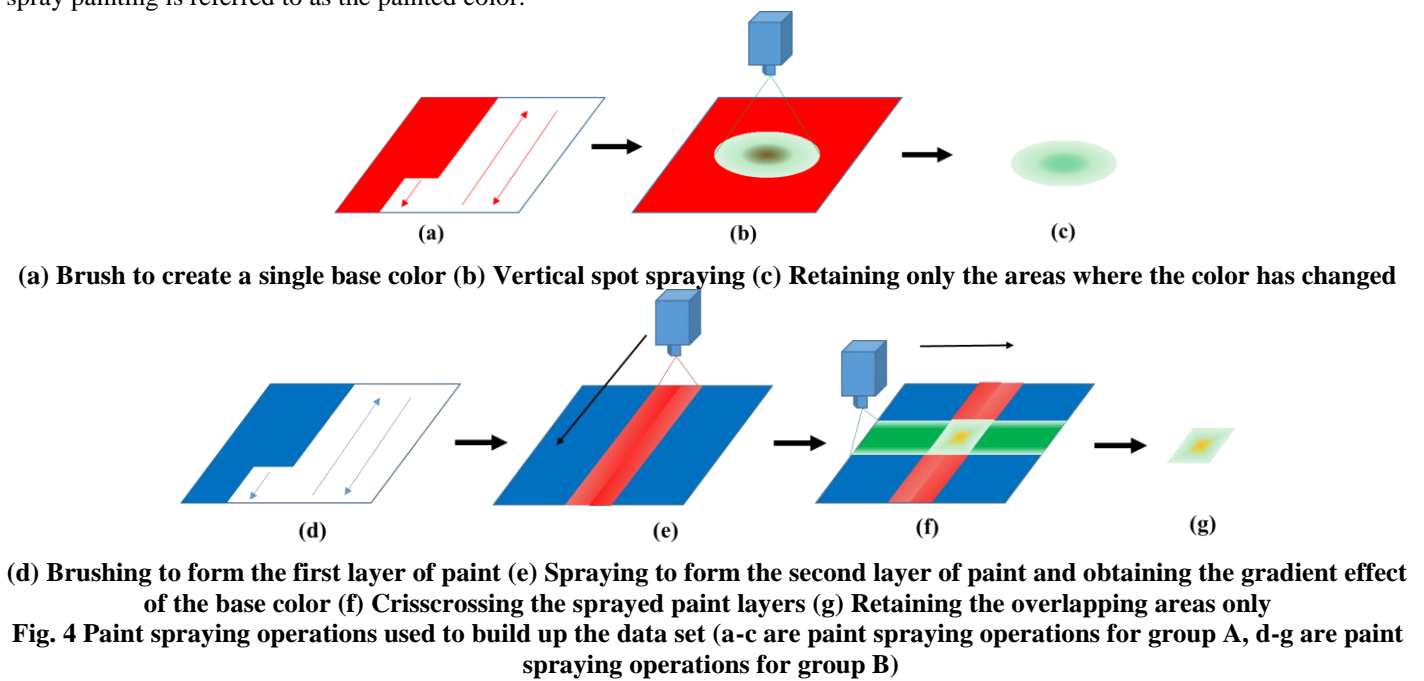

**(a) Brush to create a single base color (b) Vertical spot spraying (c) Retaining only the areas where the color has changed**

**(d) Brushing to form the first layer of paint (e) Spraying to form the second layer of paint and obtaining the gradient effect of the base color (f) Crisscrossing the sprayed paint layers (g) Retaining the overlapping areas only**

**Fig. 4 Paint spraying operations used to build up the data set (a-c are paint spraying operations for group A, d-g are paint spraying operations for group B)**

To truly show the coloring of various spraying methods in the spraying of toys and fishing gears, the following methods were followed in establishing and collecting the spray gun data. The single base color group adopts the method of uniformly applying color by brush to form a single base color for the experimental group and adopts the method of spraying at a fixed point to obtain the color after spray painting. The gradient base color group uses a first layer of paint, followed by a second layer of paint sprayed in a straight line using a robotic arm and forming a gradient base color, and finally the post-spraying color is sprayed in a cross shape using a straight-line spraying method. The schematic diagram is shown in Fig.4.

The actual painting effect during the dataset creation process is shown in Fig. 5.

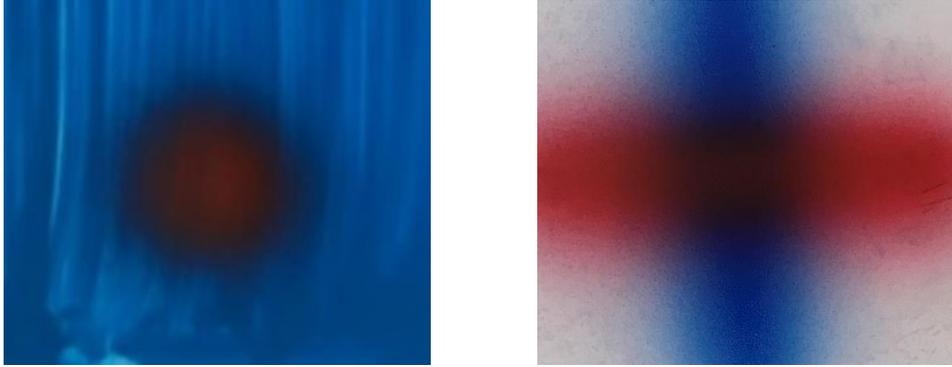

(a) Paint spraying operation of group A (b) Paint spraying operation of group B

Fig. 5 Paint spraying operation

In this paper, two groups of painting operations are carried out, A and B. Separate datasets will be created. group A is the single base color group and group B is the gradient base color group. Group A and B operations consisted of multiple sprays, respectively, and the base color, paint color and spray gun heights for each spray are shown in Table 1 and Table 2.

**Table 1 single base color group (Group A)**

| Number | Base color | Paint color | Spray gun height (cm) |
|---|---|---|---|
| 1-1 | Red | White | 12 |
| 1-2 | Red | White | 16 |
| 1-3 | Green | White | 12 |
| 1-4 | Green | White | 16 |
| 1-5 | Blue | White | 12 |
| 1-6 | Blue | White | 16 |
| 1-7 | White | Red | 12 |
| 1-8 | White | Red | 16 |
| 1-9 | Green | Red | 12 |
| 1-10 | Green | Red | 16 |
| 1-11 | Blue | Red | 12 |
| 1-12 | Blue | Red | 16 |
| 1-13 | Red | Green | 12 |
| 1-14 | Red | Green | 16 |
| 1-15 | Blue | Green | 12 |
| 1-16 | Blue | Green | 16 |
| 1-17 | White | Green | 12 |
| 1-18 | White | Green | 16 |
| 1-19 | Red | Blue | 12 |
| 1-20 | Red | Blue | 16 |
| 1-21 | Green | Blue | 12 |
| 1-22 | Green | Blue | 16 |
| 1-23 | White | Blue | 12 |
| 1-24 | White | Blue | 16 |

**Table 2 Gradient base color group (Group B)**

| Number | Base color | Paint color | Spray gun height (cm) |
|---|---|---|---|
| 2-1 | White-Red gradient | Green | 16 |
| 2-2 | White-Red gradient | Blue | 16 |
| 2-3 | White-Green gradient | Red | 16 |
| 2-4 | White-Green gradient | Blue | 16 |
| 2-5 | White-Blue gradient | Red | 16 |
| 2-6 | White-Blue gradient | Green | 16 |

## 2.3. Data collection and pre-processing

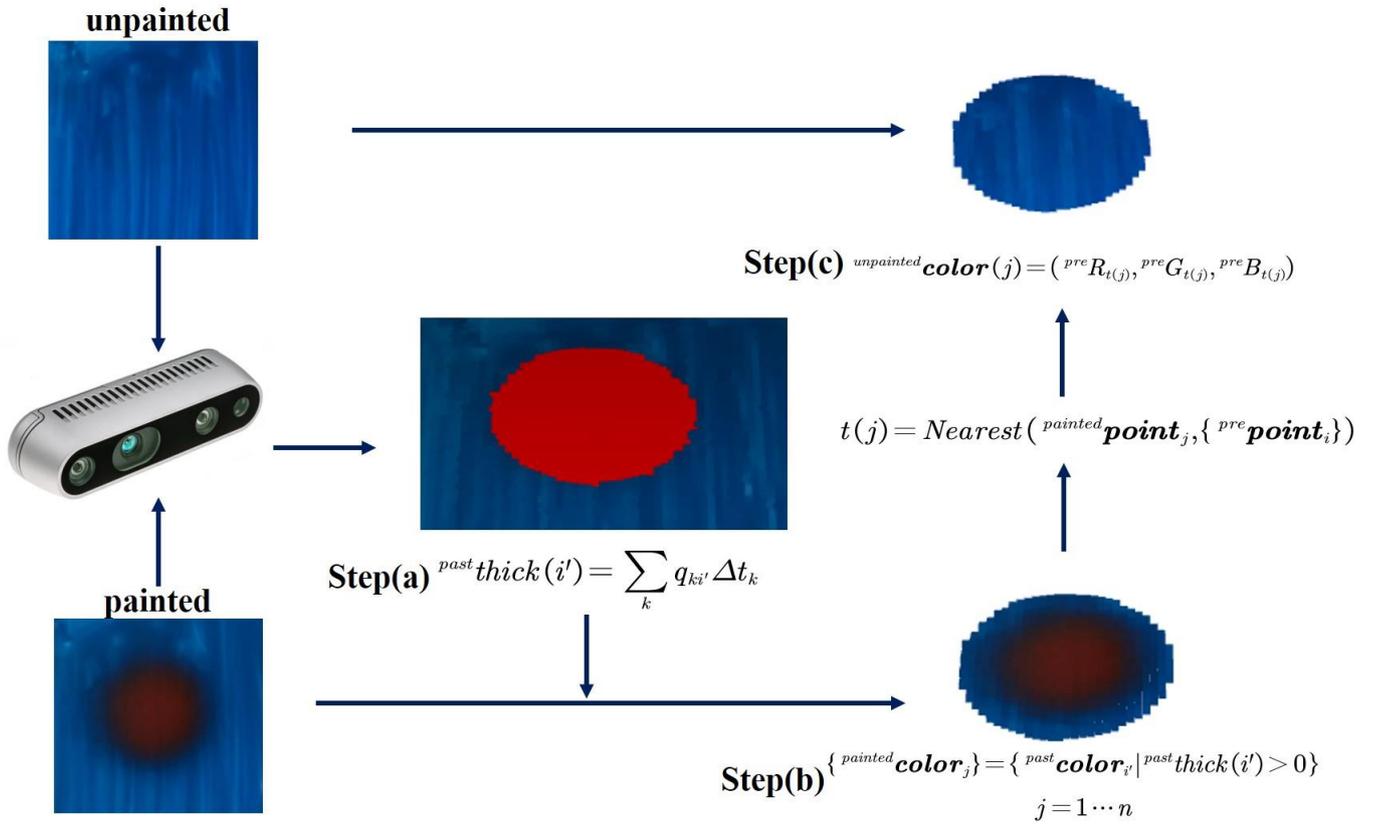

**Fig. 6** Data processing flow of group A

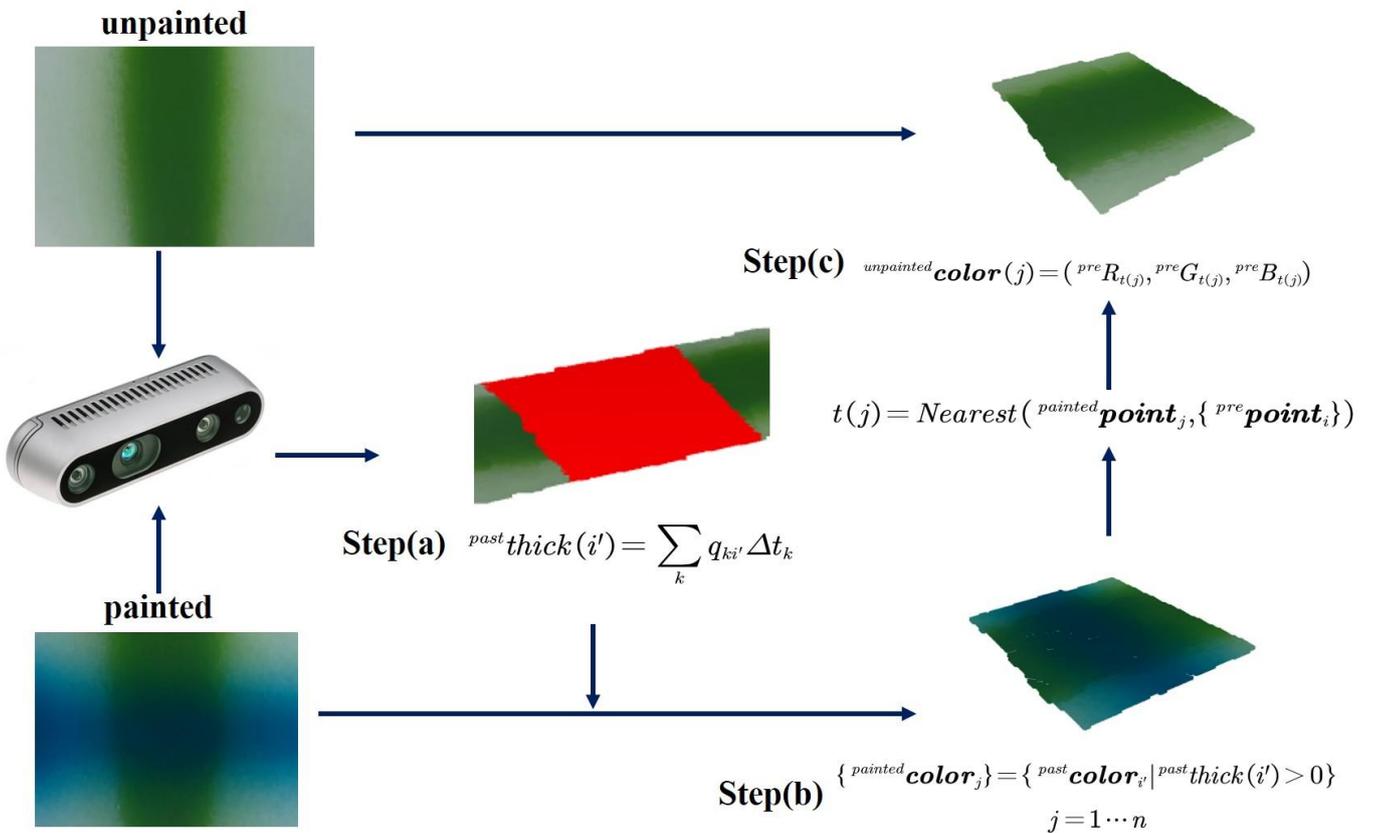

**Fig. 7** Data processing flow of group B

A color point cloud on the surface of a piece is acquired by a depth camera. Let a total of $N$ points be acquired, and the position and color of each point in the camera coordinate system are denoted as $^{pre}\boldsymbol{point}_i = (^{pre}x_i, {}^{pre}y_i, {}^{pre}z_i)$, $i = 1 \cdots N$, $^{pre}\boldsymbol{color}_i = (^{pre}R_i, {}^{pre}G_i, {}^{pre}B_i)$, $i = 1 \cdots N$. Let a total of $N'$ points be acquired on the painted surface, and the position and color of each point under the camera coordinate system are denoted as $^{past}\boldsymbol{point}_{i'}$, $i' = 1 \cdots N'$, $^{past}\boldsymbol{color}_{i'}$, $i' = 1 \cdots N'$. $N \approx N'$. The process of building the dataset for A group painting operation and B group painting operation is shown in the Fig. 6 and Fig. 7.

**Step a. Spraying Simulation**

In this paper, the first step is to simulate the spraying of the point cloud acquired by the camera. After discretizing the unpainted surface into a point cloud, the paint (simulation) thickness at each point in the point cloud can be calculated by the formula method (Eq. 5). To be able to work with the color information, the paint deposition thickness simulation is calculated directly on the unpainted point cloud acquired by the camera for the spraying trajectory of the robot (points and straight lines in this case).

The trajectory of the spray gun is a position matrix as a function of time(Eq.3)

$$Tra(t) = \begin{bmatrix} \boldsymbol{T}_{3\times3}(t) & \boldsymbol{p}_{3\times1}(t) \\ 0_{1\times3} & 1 \end{bmatrix} \tag{3}$$

Let $\boldsymbol{v}_{3\times1}(t) = \frac{d\boldsymbol{p}_{3\times1}}{dt}$, when discretizing the above equation, a strategy of discretization with respect to position is adopted. Discretize the trajectory into K trajectory points at equal intervals according to the position. Take $\|\boldsymbol{p}(k+1) - \boldsymbol{p}(k)\| = d$, $k = 1 \cdots K - 1$. Then Eq.4 is obtained.

$$Tra(k) = \begin{bmatrix} \boldsymbol{T}_{3\times3}(k) & \boldsymbol{p}_{3\times1}(k) \\ 0_{1\times3} & 1 \end{bmatrix} \tag{4}$$

Then the following equation represents the instantaneous velocity of each point $\boldsymbol{v}_{3\times1}(k) = \boldsymbol{v}_{3\times1}|_{\boldsymbol{p}(k)}$, you can calculate the equivalent residence time of each trajectory point $\Delta t_k = \frac{d}{\|\boldsymbol{v}_k\|}$.

The equivalent residence time of each trajectory point can be calculated. This model can be used in a variety of spraying simulation model, the following choice of β distribution [19]. The deposition rate of trajectory point $Tra(k)$ in $\Delta t_k$ to $^{past}\boldsymbol{point}_{i'}$, $i' = 1 \cdots N'$ is Eq.5

$$q_{ki'} = \begin{cases} A\left(1 - \frac{r_{ki'}^2}{R^2}\right)\left(\frac{H}{L_{ki'}}\right)^2 \frac{\cos\theta_{ki'}}{\cos^3\gamma_{ki'}}, & r_{ki'} \leq R, \gamma_{ki'} < \frac{\pi}{2} \\ 0 & , \text{other} \end{cases} \tag{5}$$

Where $r_{ki'} = H\tan\theta_{ki'}$, R represents the height of H when the paint can cover the maximum radius in the plane, A is the deposition rate constant, L represents the distance from the gun to the deposition point, $\theta$ represents the deposition point relative to the gun nozzle axis of the declination angle, $\gamma$ represents the angle between the normal at the deposition point (for point clouds, using normal estimation) and the line between the vent and the deposition point. This is shown in Fig.8.

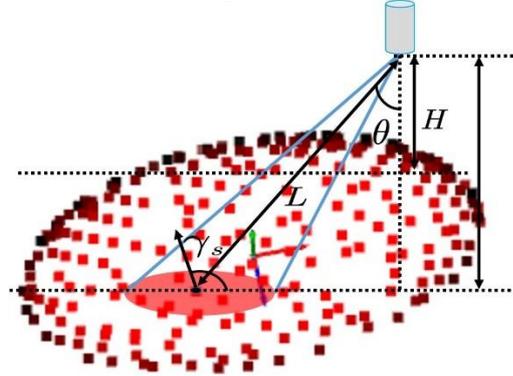

**Fig. 8 Symbolic meaning of spraying simulation formula**

The simulated paint film thickness of $^{past}\boldsymbol{point}_{i'}$, $i' = 1 \cdots N'$ is Eq.6

$$^{past}thick(i') = \sum_k q_{ki'}\Delta t_k \tag{6}$$

**Step b. Filtering point cloud data**

The machine learning in this paper only studies the part of the workpiece that is covered with a paint layer, adjusting the spraying simulation parameters so that the range of the paint coating in the simulation optimally coincides with the actual coating range, and thereafter taking the point cloud whose simulated thickness is not 0 as the point cloud covered with the paint layer.

Let there be a total number of point clouds in the study range, the data range studied by the machine learning algorithm can be expressed by Eq.7-Eq.9

$$\{^{painted}\boldsymbol{point}_j\} = \{^{past}\boldsymbol{point}_{i'}|if\ ^{past}thick(i') > 0\}, \quad j = 1 \cdots n \tag{7}$$

$$\{^{painted}\boldsymbol{color}_j\} = \{^{past}\boldsymbol{color}_{i'}|if\ ^{past}thick(i') > 0\}, \quad j = 1 \cdots n \tag{8}$$

$$\{^{unpainted}thick(j)\} = \{thick(i')|if\ thick(i') > 0\} \tag{9}$$

**Step c. Establishment of correspondence between color and simulated thickness before and after spray painting**

Let the function $Nearest(\boldsymbol{p}, \{\boldsymbol{P}_i\}) = arg \min_i \|\boldsymbol{p} - \boldsymbol{P}_i\|$, which denotes the subscript of the point in the set $\{\boldsymbol{P}_i\}$ that is closest to the point $\boldsymbol{p}$ in Euclidean distance. Take the colors of all points on the point cloud within the study area after spray painting that are closest to the points in the point cloud before spray painting as their corresponding unpainted colors as Eq.10.

$$^{unpainted}\boldsymbol{color}(j) = \left(^{pre}R_{t(j)}, ^{pre}G_{t(j)}, ^{pre}B_{t(j)}\right) \quad (10)$$
$$where \ t(j) = Nearest\left(^{painted}\boldsymbol{point}_j, \{^{pre}\boldsymbol{point}_i\}\right)$$

$class(j) = 0, 1, 2, 3 \cdots$ denotes the color code of the corresponding $^{painted}\boldsymbol{point}_j$, 0, 1, 2, 3... denotes white, red, yellow, blue... respectively.

*2.4. Data normalization and dataset segmentation*

In this paper, we first normalize $^{past}thick(i')$, $i' = 1 \cdots N'$ and use the Eq.11-E1.13

$$^{unpainted}thick_{norm}(j) = \frac{^{unpainted}thick(j) - min(^{unpainted}thick)}{max(^{unpainted}thick) - min(^{unpainted}thick)} \quad (11)$$

$$^{unpainted}\boldsymbol{color}_{norm}(j) = ^{unpainted}\boldsymbol{color}(j)/255 \quad (12)$$

$$^{painted}\boldsymbol{color}_{norm}(j) = ^{painted}\boldsymbol{color}(j)/255 \quad (13)$$

In this paper, group A and Group B are randomly divided according to the ratio of the training set: validation set: and test set = 0.9:0.05:0.05, respectively. So far, two datasets of group A and group B have been created.

## 3. Deep learning model

*3.1. Model analysis*

The purpose of this paper is to establish a regression relationship Eq.14

$$^{painted}\widehat{\boldsymbol{c}_{norm}}(j) = \boldsymbol{Model}(^{unpainted}\boldsymbol{color}_{norm}(j), ^{unpainted}thick(j), class(j)) \quad (14)$$

The significance of this is that, in the scenario of paint spraying, the color of the painted area after color spraying is predicted based on the thickness of the simulated coating on the point cloud of the workpiece, the base color of the painted area of the workpiece before spray painting, and the type of paint. When the thickness of the paint film is greater than a certain threshold, the color of the surface after color spraying is not related to the color of the surface before spray painting; when the thickness of the paint film is less than a certain threshold, the color of the surface after color spraying is closely related to the color of the surface before spray painting. The multilayer perceptron has a strong nonlinear fitting ability and good generalization ability. Therefore, this paper adopts the multilayer perceptron to establish a prediction model, and improves the multilayer perceptron model through a variety of mechanisms so that it can have a good performance in the spray painting color effect prediction problem.

*3.2. Building the model*

The classical multilayer perceptron model used to solve the spray painting color effect prediction problem is shown in Fig. 9.

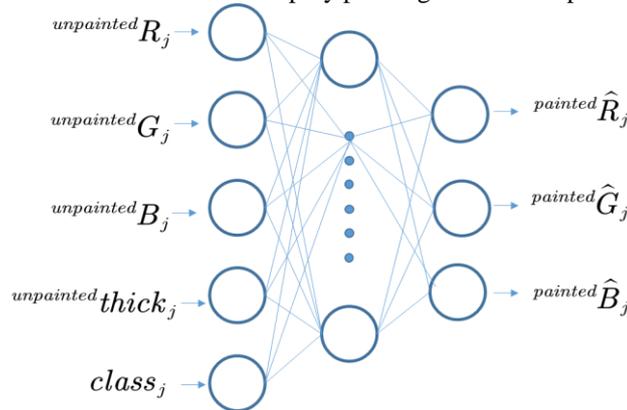

**Fig. 9 Traditional multilayer perceptron (MLP) model for modelling the problem in this paper**

Based on the illuminating knowledge from the results of the spraying operation, the base color has a greater influence on the color of the workpiece when the paint coating is smaller and thinner, and the paint color dominates the color of the workpiece when the paint coating is thicker. Therefore, the improved model employs a gating mechanism [21] to adjust the influence of the base color information on the color prediction of the workpiece according to the thickness of the paint coating.

In this paper, we hope that the deep learning model acquires as much information as possible to realize that one model can cover all workpiece products of the whole spraying production line. Therefore, the parameter size of the neural network is increased. But this leads to the problem of gradient vanishing. To solve this problem, the residual structure is added to the main prediction program in the improved model [22].

The improved multilayer perceptron model is shown in Fig.10, which is referred to as SPCP-ANN (Spray Painting Color Prediction Artificial Neural Network) in the following.

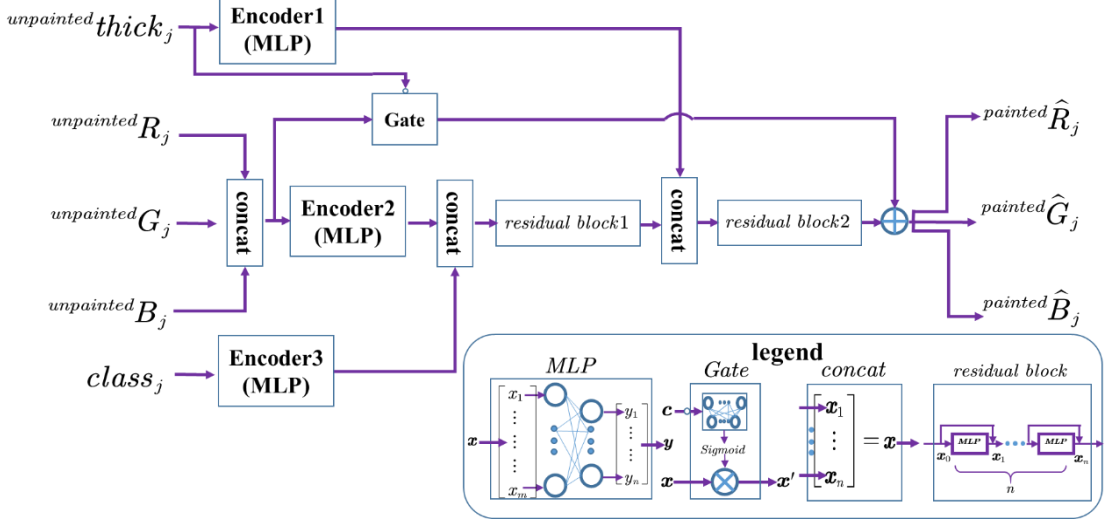

**Fig. 10 Deep learning model used for the problem in this paper (improved multilayer perceptron model, SPCP-ANN**

*3.3. Model Details*

**a. Multilayer Perceptron Module**

Let there be a total of N layers (including output layers) in a multilayer perceptron module when the input layer is not counted (set to layer 0). For layer $l \in \{1, \cdots, N\}$, let layer l-1 have m neurons (i.e., layer l has m input variables) and layer l has n neurons (i.e., layer l has n output variables). The relationship between input variables and output variables is satisfied as Eq.15

$$^l x_{n \times 1} = f(W_{n \times m}{}^{l-1} x_{m \times 1} + b_{n \times 1}) \tag{15}$$

In this paper, except for the multilayer perceptron in the gating module, the activation function used is uniform as Eq.16

$$f(x_{n \times 1}) = [relu(x_1), \cdots, relu(x_n)]^T \text{ where } relu(x) = \begin{cases} x, & x \geq 0 \\ 0, & x < 0 \end{cases} \tag{16}$$

The activation function used in the multilayer perceptron for the gating mechanism is Eq.17

$$sigmoid(x) = \frac{1}{1 + e^{-x}} \tag{17}$$

A multilayer perceptual machine with m input variables, n output variables, the number of layers is N (the number of neurons per layer is set separately), and the activation functions are all can be represented as Eq.18

$$y_{n \times 1} = MLP(m, n, N, f, x_{m \times 1}) \tag{18}$$

**b. Gate Control Module**

Define the symbols ⊗ to denote the multiplication of corresponding elements of a vector or matrix. The input that acts as a control is $c$, and the controlled input variable is $x$. The relationship between the input variable and the output variable is Eq.19

$$x_{n \times 1}' = x_{n \times 1} \otimes MLP(m, n, N, f = sigmoid, c_{m \times 1}) \tag{19}$$

**c. Vector Link Module**

Input n vectors $\{x_{m_1 \times 1}, \cdots, x_{m_n \times 1}\}$ and concatenate them into a new vector as Eq.20

$$x_{\sum m_i \times 1} = [x_{m_1 \times 1}{}^T, \cdots, x_{m_n \times 1}{}^T]^T \tag{20}$$

**d. Residual module**

Let there are n improved multilayer perceptron modules with residual structure between inputs and outputs in the residual module, then the relationship between the input $x_{i-1}$ of the ith module and the output $x_i$ of the ith module is EQ.21

$$x_i = MLP(m_i, n_i, N_i, f_i, x_{i-1}) + x_{i-1} \tag{21}$$

**e. Loss Functions and Parameter Optimizers during Training**

Let there be a total of n data points $\{x_i, y_i | i = 1 \cdots n\}$ in the training set, which is divided into several batches of size m. The predicted value of each data point of the model is $\hat{y}_i$. Denoted $\|y\|$ by the L2 norm of $y$, the goal of model training is to minimize the loss function (Eq.22)

$$Loss = \frac{1}{n} \sum_{i=1}^{n} \|y_i - \hat{y}_i\|^2 \tag{22}$$

In actual training, each time the parameters are optimized, the optimization objective is to minimize the loss function (Eq.23) of a batch of data

$$Loss' = \frac{1}{m}\sum_{j=1}^{m} \|y_j - \hat{y}_j\|^2 \tag{23}$$

During training, the parameters are optimized using the Adam optimizer [20].

## 4. Experiments and results

*4.1. Comparison of the effects of deep learning methods and interpolation methods*

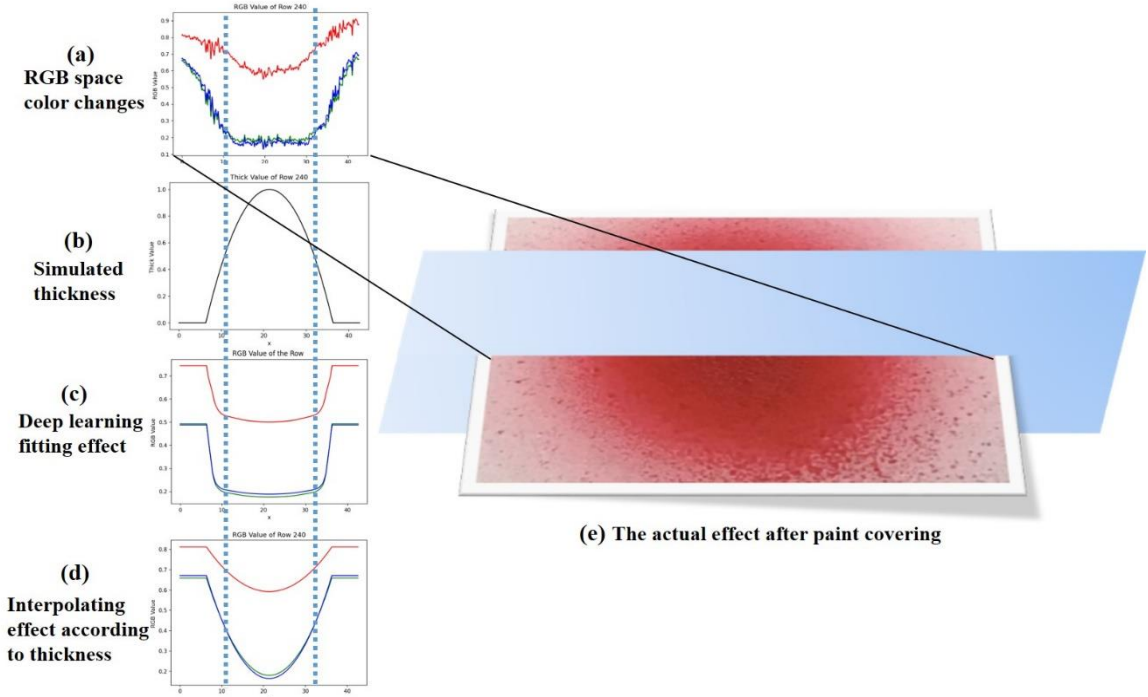

**Fig. 11 Ability of deep learning to predict the color effect of spray painting**

The advantage of the deep learning method over color gradation based on simulated thickness is that it solves the problem that the rate of change of simulated thickness is different from the rate of change of actual thickness. The advantage over color gradation directly for RGB is that the gradation region is determined automatically. Fig.11 (e) shows the actual effect of a certain spray paint coverage, (a) (b) shows the RGB space color change of the cross-section and the simulated thickness of the paint film, (c) is the RGB space color of the cross-section predicted using the deep learning method, and (d) is the RGB space color of the cross-section predicted using the linear interpolation method with the thickness as the independent variable (the RGB is the dependent variable, respectively). It can be seen that the deep learning method has a significant advantage in determining the gradient region and gradient rate.

*4.2. Comparison of color prediction effect and real effect*

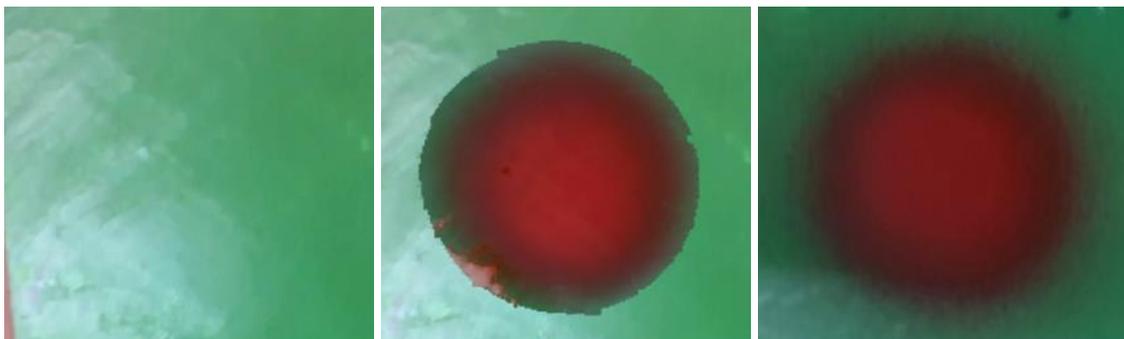

**(a)Green base color (b) Predicted effect of red paint spraying under green base color (c) Actual effect of red paint spraying under green base color**

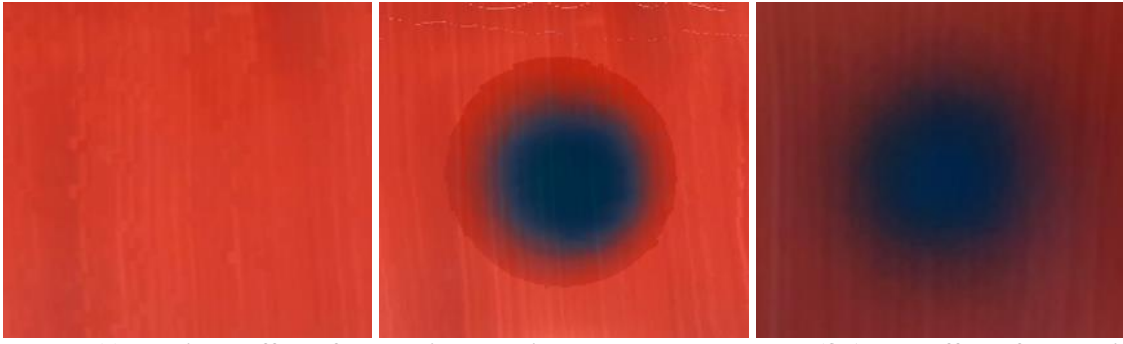
**(d) Red base color (e) Predicted effect of blue paint spraying under red base color (f) Actual effect of blue paint spraying under red base color**

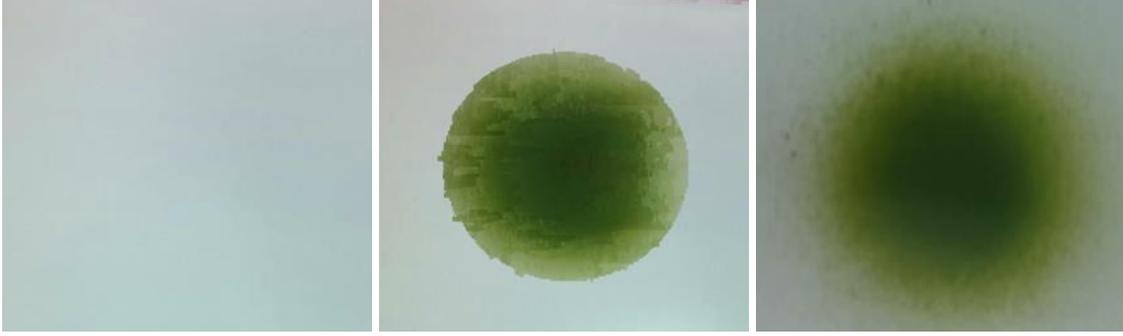
**(g) White base color (h) Predicted effect of green paint spraying under white base color (i) Actual effect of green paint spraying under white base color**
**Fig. 12 Comparison between predicted and actual effects**

As shown in Fig.12, in terms of viewing, the spray painting color effect prediction model established in this paper can effectively predict the color effect after spray painting.

### 4.3. Training and Testing Analysis of Deep Learning Models

To demonstrate the advantages of the improved multilayer perceptron model (SPCP-ANN) proposed in this paper on the paint color prediction task of spraying robots, two neural networks were constructed as the baseline models for comparative testing. One of them is the PSO single-hidden-layer MLP model, which is a single hidden layer (i.e., a total of two layers of neurons) multilayer perceptron after optimizing the hyperparameters with the particle swarm algorithm (it can be considered as an optimal kind of structure of the classical single hidden layer multilayer perceptron for the task of this paper). The significance of the comparison with this model is to illustrate the performance advantage of the present model over the classical multilayer perceptron. The multi-hidden-layer MLP, on the other hand, is a multilayer perceptron with multiple hidden layers. The significance of the comparison with this model is to compare the performance advantage of this model over the multi-hidden layer perceptron model with the same number of parameters. The training consumption of the above three models is shown in Table 3. The root means square error (RMSE) and means relative error (MRE) are given as Eq.24 and Eq.25

$$RMSE = \sqrt{\frac{1}{n}\sum_{i=j}^{n} \|{}^{painted}\widehat{c_{norm}}(j) - {}^{painted}color_{norm}(j)\|^2} \tag{24}$$

$$MRE = \frac{1}{n}\sum_{j=1}^{n} \frac{\|{}^{painted}\widehat{c_{norm}}(j) - {}^{painted}color_{norm}(j)\|}{\|{}^{painted}color_{norm}(j)\|} \times 100\% \tag{25}$$

**Table 3 Comparison of training consumption of the proposed model in this paper with the baseline model(1k=1000)**

| Training Consumption | PSO Single-Hidden-Layer MLP | Multi-hidden-Layer MLP | Improved MLP (Ours SPCP-ANN) |
|---|---|---|---|
| Optimizer | Adam | Adam | Adam |
| **Learning rate** | 0.0087 (after PSO optimization) | 0.001 | 0.001 |
| Number of parameters | ≈1k (after PSO optimization) | ≈155k | ≈155k |

The changes in RMSE of the validation set of the above three perception models during the training process are shown in Fig. 13. The improved MLP model presented in this paper is better trained at epoch 30 than the two baseline models at epoch 30 and later.

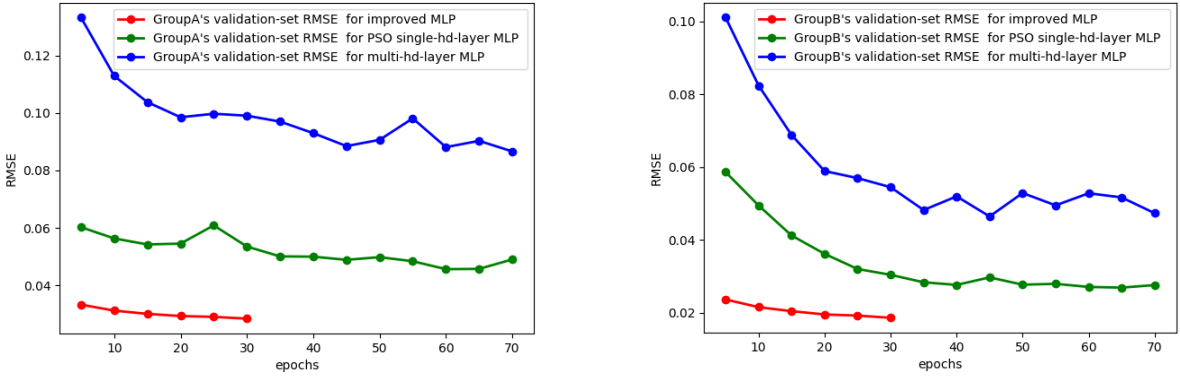

**Fig. 13 Variation of validation set RMSE with training on group A (left), B (right) dataset**

A comparison of the test set evaluation metrics of the above three models in terms of RMSE and MRE is shown in Table 4.

**Table 4 Comparison of test set metrics between the model proposed in this paper and the baseline model on group A, and B dataset**

| Dataset | Medel | RMSE | MRE |
|---|---|---|---|
| A | Multi-hidden-layer MLP | 0.0840 | 21.10% |
| | Single-hidden-layer MLP with PSO | 0.0464 | 11.64% |
| | **Ours (SPCP-ANN)** | **0.0288** | **7.743%** |
| B | Multi-Hidden-Layer MLP | 0.0584 | 28.01% |
| | Single-Hidden-Layer MLP with PSO | 0.0258 | 12.35% |
| | **Ours (SPCP-ANN)** | **0.0177** | **6.814%** |

According to Table 4, the model SPCP-ANN proposed in this paper has better performance relative to both PSO-optimized (for optimal hyperparameters) single-hidden-layer MLP and multi-hidden-layer MLP as indicated by the bolded portion. This shows the effectiveness of improved methods (gate control blocks and residual blocks) in spray painting color effect prediction for multi-layer perceptron (MLP) model and also verifies that the model proposed in this paper can effectively predict spray painting color effect.

*4.4. Comparison of Deep Learning Models with Other Machine Learning Models*

**Table 5 Some key hyperparameter settings for machine learning models**

| Algorithms | Adaboost | RF | SVR | LR |
|---|---|---|---|---|
| Hyper parameterization | Base estimator: decision tree<br>Number of estimators: 50<br>Learning rate: 1<br>Objective function: MSE | Number of decision trees: 100<br>Maximum depth: 8 layers<br><br>Objective function: MSE | Kernel function: RBF<br>Penalty parameter C: 1.0<br>Gamma: 0.2<br>Objective function: MSE | — |

**Table 6 Test set RMSE and MRE of each model on group A, B dataset**

| Dataset | Model | RMSE | MRE |
|---|---|---|---|
| A | LR | 0.1655 | 42.65% |
| | Adaboost | 0.0942 | 23.20% |
| | SVR | 0.0597 | 15.39% |
| | Random Forest | 0.0476 | 12.30% |
| | **Ours (SPCP-ANN)** | **0.0288** | **7.743%** |
| B | LR | 0.0753 | 36.28% |
| | Adaboost | 0.0509 | 24.62% |
| | SVR | 0.0498 | 23.84% |
| | Random Forest | 0.0228 | 10.93% |
| | **Ours (SPCP-ANN)** | **0.0177** | **6.814%** |

To reflect the performance advantage of the improved MLP in this paper over the traditional machine learning model on the paint color prediction task of spraying robots, the performance of Support Vector Machine Regression (SVR) [23], Random Forest (RF) regression [24], Adaboost [25], and Linear Regression (LR) models were tested on two datasets, A and B, respectively. The parameter settings of each algorithm are shown in Table 5. The performance metrics of each machine learning model on both datasets A, and B are shown in Table 6.

According to the bold part of Table 6, the performance of the model proposed in this paper is significantly better than that of traditional machine learning methods on the spray painting color effect prediction problem of spraying robots. This is not only because the deep learning method has more parameters, can carry more information, and has stronger nonlinear fitting ability, but also because this paper improves the multilayer perceptron according to the characteristics of the problem.

*4.5. Practical application cases*

To demonstrate the practical value of the model, this paper will show its application in gun trajectory optimization. The main tasks to be achieved in this section are as follows: given a sample with only partial spraying and an initial trajectory that does not match the spraying area of the sample, the trajectory that matches the spraying area of the sample is obtained by optimizing the initial trajectory.

Spraying experiments are carried out on a lure bait to establish a data set and train the prediction model. The dataset is built as before, see Fig. 14. The prediction of the spray painting color effect on the lure bait is shown in Fig. 15.

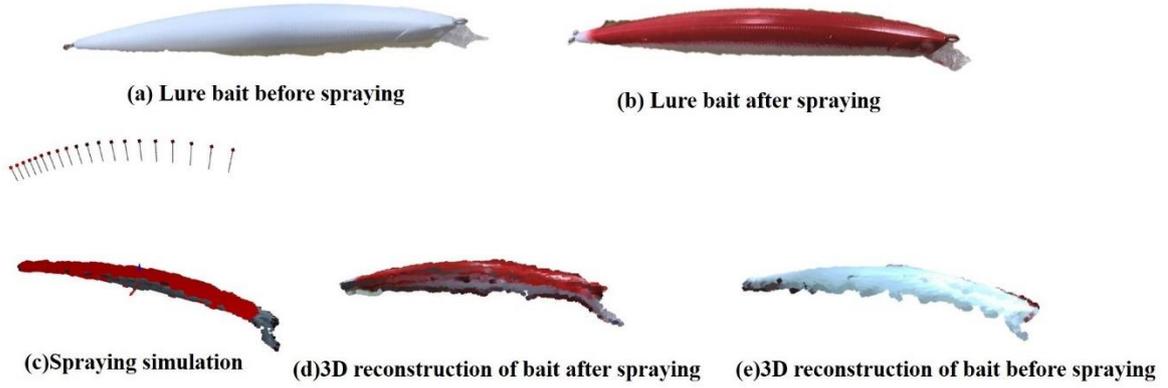

(a) Lure bait before spraying  (b) Lure bait after spraying

(c) Spraying simulation  (d) 3D reconstruction of bait after spraying  (e) 3D reconstruction of bait before spraying

**Fig.14 Collecting data and training the model on a lure bait**

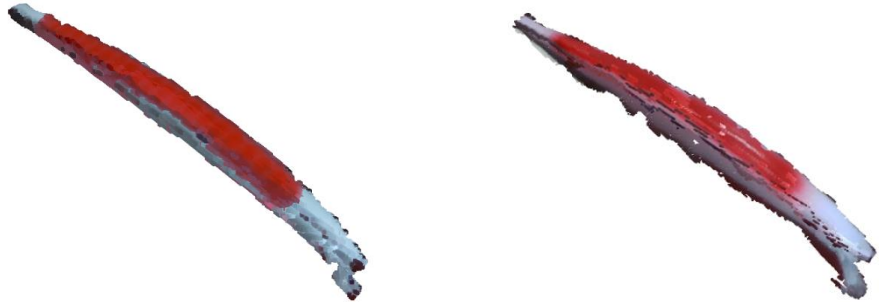

(a) Predicted effect of spray paint color on lure bait    (b) Real effect of spray paint color on lure bait
**Fig. 15 Comparison between the predicted and real effects of the SPCP-ANN model on lure baits**

$$minimize \quad f_1(x), f_2(x) \quad (26)$$

$$subject\ to \begin{cases} H_{min} < H_k < H_{max}, k = 1, 2, \cdots, K \\ v_{min} < v_k < v_{max} \\ 0 \leq conf_k \leq 1 \end{cases}$$

$$where \quad x = [H_1 \cdots H_k \ v_1 \cdots v_k \ conf_1 \cdots conf_K]$$

$$f_1(x) = \sqrt{\frac{1}{n}\sum_j \|(\widehat{color}(j) - {}^{sample}color(j))\|^2}, f_2(x) = \sum_k \Delta t_k$$

This paper demonstrates the potential of the SPCP-ANN model for paint trajectory optimization through a practical example. An optimization model, Eq. 26, is developed.

Point cloud slicing method [26] is used to obtain the initial trajectory. The objective function $f_1(x)$ indicates that the predicted painted color of the bait surface is as close as possible to the sample color, and $f_2(x)$ indicates that it takes as little time as possible. Where $H_{ij}\ v_{ij}\ confi_k$ denotes the height (distance from the lure bait) of the trajectory control point, the speed (the height and speed of the other trajectories can be obtained by interpolation), and the spraying confidence (the spraying confidence of the other trajectory points can be obtained by interpolation, and the spray gun will only be turned on at the trajectory points with the confidence greater than or equal to 0.5), respectively.

For the multi-objective optimization model, it can be solved quickly using NSGA-II. NSGA-II (Non-dominated Sorting Genetic Algorithm II) [27] is a multi-objective optimization algorithm for solving optimization problems with multiple objective functions.

It combines the ideas of Genetic Algorithm and Non-dominated Sorting and aims to find a set of Pareto optimal solutions, i.e., solutions that cannot be dominated by other solutions. The Pareto frontier graph of the solution results is shown in Fig. 16.

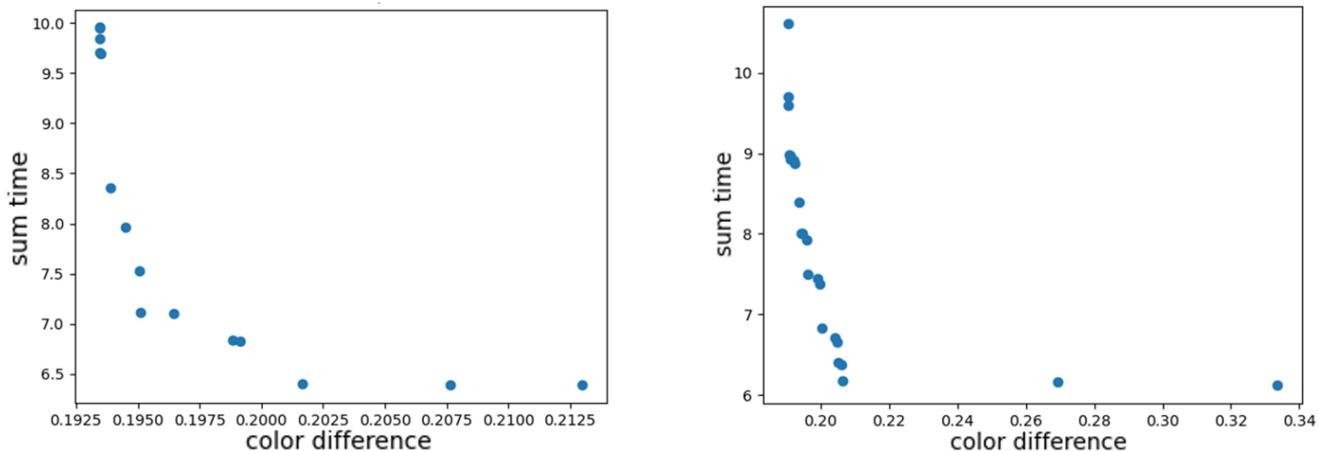

(a) Pareto front plot of solution results for optimization case 1    (b) Pareto front plot of solution results for optimization case 2

**Fig. 16 Pareto front plot of the solution derived by NSGA-II**

By optimizing the effect (compare Fig 17(b) and (d), or compare Fig 18(b) and (d)), it can be seen that the trajectory is optimized from not being properly painted locally to being able to paint a specific area correctly. It can be seen that the method in this paper can effectively replace the simulated thickness and solve the trajectory planning problem of local spraying in multi-color spraying.

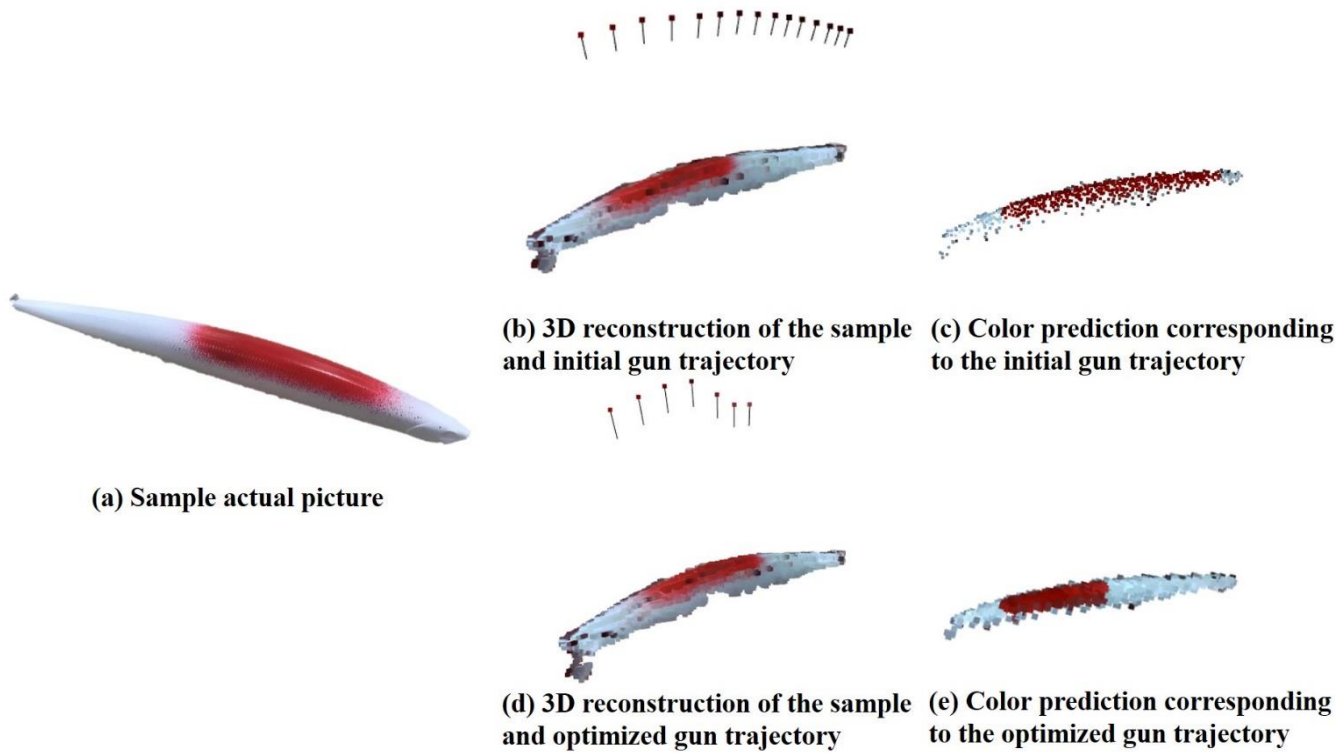

**Fig. 17 Optimization effect of optimization case 1**

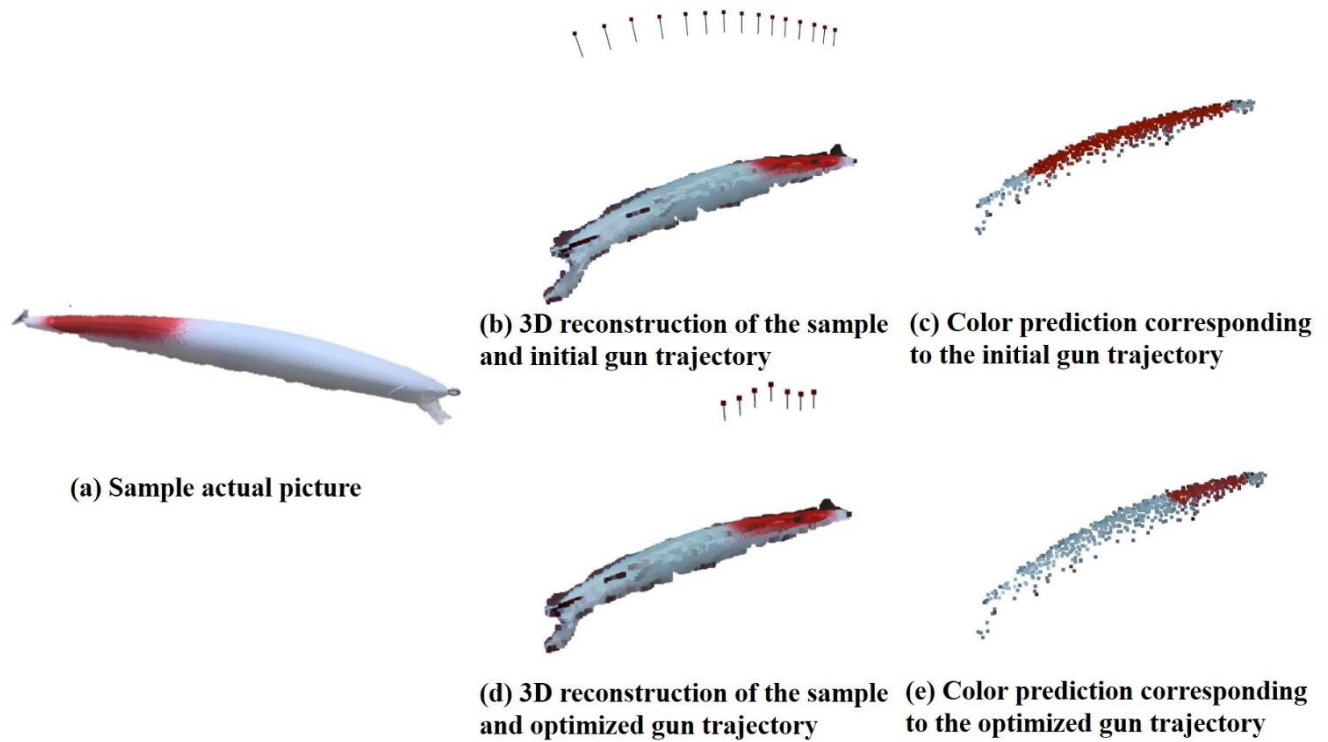

Fig.18 Optimization effect of optimization case 2

## 5. Conclusion

In this paper, we propose an improved multilayer perceptron inspired by the Kubelka-Munk model, the Spray Paingting Color Prediction Artificial Neural Networks (SPCP-ANN). The SPCP-ANN achieves pixel-level prediction of spray painting color effects for the spray gun trajectory optimization problem of a spray-painting robot. In this paper, we detail the method of building the dataset, verify the performance of SPCP-ANN through comparative experiments, and give a real example of spray gun trajectory optimization by applying the model. The comparison experiments show that the mean relative errors of SPCP-ANN are 7.743% and 6.814% on the datasets of single base color and gradient base color, respectively, which are significantly better than the baseline model. Through the example, this paper shows the potential of the SPCP-ANN in spray gun trajectory optimization for spray-painting robots. The main significance of this paper is that a new method for constructing the objective function of gun trajectory optimization is given, which takes the difference between the predicted surface color and the sample surface color of workpiece as the optimization objective. While the conventional method takes the difference between the simulated thickness and the desired thickness as the optimization objective, this new gun trajectory optimization objective can solve the problems of the inability to optimize the trajectory for multi-color spraying as well as the difficulty in controlling the color gradient region caused by the traditional method (due to the difficulty in inverting the thickness from the color). The work done in this paper can help the promotion of automatic spraying technology in multi-color spraying scenarios.

**Acknowledgments**

This research is supported by the National Natural Science Foundation of China (Grant No. 52475597), the Natural Science Foundation of Shandong Province (ZR2022ME084), and the Young Scholars Program of Shandong University, Weihai.